# Advancements in Content-Based Image Retrieval: A Comprehensive Survey of Relevance Feedback Techniques

Hamed Qazanfari
*Image Processing and Data Mining Lab, Shahrood University of Technology,
Shahrood, Iran*
hamedmit.gh@gmail.com

Mohammad M. AlyanNezhadi
*Faculty of Science, University of Science and Technology of Mazandaran, Behshahr, Iran*
alyan.nezhadi@mazust.ac.ir

Zohreh Nozari Khoshdaregi
*Department of Computer Engineering University of Bojnord, Iran*
z.nozari222@gmail.com

*Abstract*— Content-based image retrieval (CBIR) systems have emerged as crucial tools in the field of computer vision, allowing for image search based on visual content rather than relying solely on metadata. This survey paper presents a comprehensive overview of CBIR, emphasizing its role in object detection and its potential to identify and retrieve visually similar images based on content features. Challenges faced by CBIR systems, including the semantic gap and scalability, are discussed, along with potential solutions. It elaborates on the semantic gap, which arises from the disparity between low-level features and high-level semantic concepts, and explores approaches to bridge this gap. One notable solution is the integration of relevance feedback (RF), empowering users to provide feedback on retrieved images and refine search results iteratively. The survey encompasses long-term and short-term learning approaches that leverage RF for enhanced CBIR accuracy and relevance. These methods focus on weight optimization and the utilization of active learning algorithms to select samples for training classifiers. Furthermore, the paper investigates machine learning techniques and the utilization of deep learning and convolutional neural networks to enhance CBIR performance. This survey paper plays a significant role in advancing the understanding of CBIR and RF techniques. It guides researchers and practitioners in comprehending existing methodologies, challenges, and potential solutions while fostering knowledge dissemination and identifying research gaps. By addressing future research directions, it sets the stage for advancements in CBIR that will enhance retrieval accuracy, usability, and effectiveness in various application domains.

## I. Introduction

Content-based image retrieval (CBIR) systems play an important role in the field of computer vision. These systems search for images based on their visual content rather than metadata, such as keywords and tags. While CBIR focuses on image retrieval, it can be closely related to the broader topic of object detection [1]. Object detection involves identifying and localizing specific objects within an image [2]. This automated approach allows for the retrieval of images that share visual similarities with the query image. Overall, content-based image retrieval approaches enhance the object detection process by enabling the identification and retrieval of visually similar images based on their content features [1].

In CBIR systems, low-level features like color, texture, and shape are extracted to describe the content of images. These features can be utilized in object detection to detect and classify objects based on their visual characteristics [3].

Challenges in CBIR include the semantic gap, where low-level visual features don't directly correspond to high-level semantic concepts humans use to interpret images. Bridging this gap is crucial to improve CBIR accuracy and relevance [4]. One of the solutions to address these challenges is Relevance Feedback (RF). RF allows users to provide feedback on the retrieved images, indicating their preferences and iteratively refining the search results. Incorporating RF into CBIR systems enhances accuracy by adapting to user preferences and reducing the semantic gap [4].

In general, CBIR is a vital component of object detection and computer vision, and providing a survey on CBIR and RF is crucial. A comprehensive survey helps researchers and practitioners understand the existing techniques, challenges, and potential solutions. It facilitates knowledge sharing, identifies research gaps, and paves the way for advancements in CBIR and RF in the future.

The rest of this paper is organized as follows. We discuss the CBIR and The RF in more detail in the next sections. Then in section 2 the categories of the RF method and some of the related works will be discussed. The paper concludes in Section III with a discussion of the results achieved and some suggestions for future work.

### A. CBIR Systems

Content-Based Image Retrieval (CBIR) is a crucial component of computer vision, enabling the retrieval of images based on their visual content rather than relying on metadata or textual descriptions. CBIR has various applications, including object detection, image recognition, and multimedia retrieval [5,6]. However, CBIR faces several challenges that affect its performance and usability. Here is an overview of the CBIR system, its challenges, and potential solutions:

CBIR involves feature extraction, where low-level visual features are extracted from images. These features are used to build a representation of the image content. CBIR algorithms compare these feature representations to retrieve similar images from a database. The retrieval can be based on a query image or specified visual features. CBIR algorithms compare the query image or features with the feature representations of the images in the database to retrieve the most similar ones [5,6].

1. Challenges in CBIR:

-Semantic Gap: The semantic gap refers to the difference between low-level visual features and high-level semantic concepts that humans use to interpret images [4,7]. Bridging this gap is challenging, as CBIR focuses on extracting and comparing low-level features, which may not directly correspond to the intended semantic meaning.

- Scalability: As the volume of images in the database increases, the efficiency and effectiveness of CBIR systems may decrease [4,7]. Searching through a large number of images can become computationally expensive and time-consuming.

2. Solutions:

- Utilizing a combination of low-level features: These methods employ both low-level features and generate additional semantic features through them. One example is *CDHTxEo* (Colour Difference Histogram of Image Texture and Edge Orientation), which goes beyond a basic histogram to extract image semantics. We have presented several papers falling into this category [1,3,8]. These approaches do not require vast amounts of data or extensive training, yet they can yield a CBIR system with satisfactory retrieval rates and improved mean average precision.

- Machine Learning Techniques: Various types of machine learning algorithms are used for training and extracting low-level features and high-level semantics of the images respectively. In our study mentioned in [9], we introduced a robust SVM algorithm to train our newly proposed features. Recent advancements in deep learning, particularly deep Convolutional Neural Networks (CNNs), have shown promise in addressing the challenges of CBIR [4,10]. CNNs can extract more meaningful and higher-level features from images, helping to bridge the semantic gap and improve retrieval accuracy [4].

- Relevance Feedback (RF): RF enables users to provide feedback on the retrieved images, which is then used to refine the search results and improve relevance [11]. Incorporating RF into CBIR systems helps adapt the search process to user preferences and reduces the semantic gap [11,12]. In [11], we introduced a short-term learning algorithm for CBIR based on RF. This algorithm will be thoroughly explained in the following section.

In conclusion, CBIR is a critical component of computer vision, enabling image retrieval based on visual content. Nevertheless, challenges such as the semantic gap and scalability exist. Advances in deep learning techniques, particularly CNNs, provide approaches to address these challenges. Additionally, integrating RF into CBIR systems allows for user feedback and iterative refinement of search results, enhancing the relevance and performance of the retrieval process.

*B. Relevance Feedback*

Relevance Feedback (RF) is a technique used in CBIR systems to improve the accuracy and effectiveness of image retrieval. RF allows users to provide feedback on the relevance of retrieved images, which helps refine subsequent search iterations [13]. This response will provide a comprehensive explanation of RF in CBIR, including its methodology, challenges, and two categories within RF [11]: Short-Term Learning (STL) and Long-Term Learning (LTL).

RF in CBIR involves a feedback loop between the user and the system, typically consisting of three main steps: query formation, result analysis, and feature space adaptation. Initially, the user submits a query to retrieve a set of images. After viewing the retrieved results, the user provides feedback by marking images as relevant or irrelevant to their search. This feedback is then utilized to iteratively refine the query or adjust the feature weights, enhancing subsequent search results.

The RF method faces several challenges in CBIR systems. One significant challenge is the subjectivity and inconsistency of user feedback. Different users have diverse interpretations of relevance, leading to variations in feedback. Overcoming this challenge requires techniques to interpret and adapt to user preferences effectively [14].

Because RF in CBIR is divided into two subcategories, STL and LTL, we conducted a survey on these. In the remainder of the paper, we provided a brief explanation of some new research from these two categories.

II. RESEARCH BACKGROUND

Within the framework of RF, two main categories have emerged: Short-Term Learning (STL) and Long-Term Learning (LTL).

1. Short-Term Learning (STL):

STL methods focus on immediate adaptation based on user feedback within a single session of retrieval. The system dynamically adjusts the feature representation or query based on this feedback. Common techniques in STL include query expansion, query modification, and result re-ranking [15].

- Query expansion involves expanding the original query by adding additional terms extracted from relevant images. The expanded query aims to capture more relevant images during subsequent retrieval iterations.

- Query modification modifies the original query based on user feedback, either by removing irrelevant terms or incorporating new terms based on the relevant images.

- Result re-ranking adjusts the ranking order of initially retrieved images based on feedback. Images marked as relevant receive higher priority in subsequent search iterations.

- Combination of methods, means using two or more methods that were mentioned before.

2. Long-Term Learning (LTL):

LTL methods focus on cumulative learning across multiple user sessions. The system learns from feedback provided by a larger user community over an extended period. LTL approaches utilize this aggregated feedback to refine and adapt the retrieval model for subsequent users. Techniques such as collaborative filtering, user clustering, and personalized models are common in LTL [16,17]. LTL and STL can be used together in a CBIR system [18]. They work in harmony to enhance both the learning capability over time and the accuracy of each query.

- Collaborative filtering analyzes the feedback patterns and preferences of a user community to identify relevant images for individual users. It leverages the collective wisdom of the community to improve individual search results.

- User clustering categorizes users with similar preferences into groups. The system then learns from the feedback of similar users to personalize the retrieval model for individuals within the same cluster.

- Personalized models create unique user profiles based on feedback history. These models capture individual preferences and adaptation patterns to enhance the retrieval process for each user.

By incorporating user feedback through relevance feedback, the proposed method adapts the CBIR system to user preferences and improves the overall retrieval accuracy. The subjective human perception of image similarity is leveraged to enhance the system's performance according to the user's viewpoint [11].

In CBIR, the short-term relevance feedback algorithm is more commonly used. This is because it focuses on immediate adjustments to the retrieval process based on user feedback. The short-term approach allows users to provide feedback on specific queries, refining the search results for the current session. It offers a quick and iterative process, making it suitable for real-time retrieval scenarios. While long-term relevance feedback algorithms provide sustained improvements over time, they require more extensive data collection and analysis. The short-term approach is more practical for regular users who require quick and relevant results without the need for extensive feedback history [19].

*A. Long-term Learning*

The authors of [20] introduce an improved method for improving the accuracy of CBIR by incorporating user feedback. They use feature re-weighting, classification techniques, as well as a new idea to enhance the accuracy of the initial retrieval. The study illustrates how deep learning features and clustering methods result in significant performance improvements.

A key innovation of this paper is the introduction of grouping relevant images. After the initial retrieval using the classical CBIR approach, similar or relevant images are grouped together. This grouping helps to organize and structure the relevant images, providing better understanding of their characteristics. The grouping process continues iteratively as users provide feedback. New groups are formed by adding relevant images to existing groups or creating new ones. This iterative grouping approach leverages feedback information to refine the grouping structure and improve retrieval accuracy. The proposed method shows that utilizing the grouping information greatly enhances retrieval accuracy. The grouped images capture shared characteristics or semantic relationships among relevant images. This information helps refine the ranking criteria and retrieve more relevant images in subsequent iterations. Another significant aspect of the proposed method is the improvement in retrieval accuracy even in the initial retrieval attempt. Usually, the initial results in the first attempt may not be perfect. However, by incorporating relevance feedback from users and leveraging the grouping structure, the proposed method achieves enhanced accuracy even from the first retrieval attempt [20].

The proposed method in [16] is called Case-based Long-Term Learning (CB-LTL), which is designed for CBIR systems. CB-LTL combines Case-Based Reasoning (CBR) and Long-Term Learning (LTL) to enhance the retrieval performance of CBIR systems across multiple query sessions.

CB-LTL consists of two main stages: the learning stage and the reasoning stage. In the learning stage, the system creates and updates a Case Knowledge Base (CKB) by recording relevant cases from retrieval sessions. Each case in the CKB includes the user's query and the system's retrieved images. The information of each case is stored in a structure called Semantic Frame (SF), which contains low-level features, high-level features, and the frequency of visitation of that case [16].

During the reasoning stage, the SFs in the CKB are used to improve the retrieval results of the current session. A trigger function is employed to search for similar cases to the current session. If a similar case is found, its SF is updated based on the frequency of visitation. If no similar case is found, the current session is added as a new case to the CKB. The CB-LTL method makes use of the stored information in the SFs to enhance the retrieval performance. It utilizes the low-level features to refine the query and the high-level features to improve the retrieval results. The SFs also assist in initializing or improving the Short-Term Learning (STL) process by providing meta-information like similarity functions or classifiers [16].

The CB-LTL method has been implemented and evaluated on various datasets, demonstrating promising results in enhancing retrieval precision compared to CBIR systems without LTL. Moreover, it remains robust against noise and can track progress over multiple query sessions [16].

Baig et al. [21] present a method for content-based image retrieval using SURF (speeded-up robust features) and CoHOG (co-occurrence histograms of oriented gradients) descriptors integrated and relevance feedback into a bag-of-visual-words (BoVW) model to address semantic gap and overfitting issues. It utilizes a weighted polling scheme, linear discriminant analysis (LDA), and relevance feedback to enhance image retrieval accuracy and performance.

The purpose of the RF in the proposed method is to improve the search strategy of image retrieval iteration by iteration, refining image retrieval results based on user-assigned labels of relevant and irrelevant images as positive and negative feedback samples. It aims to learn user preferences in each iteration and improve the classification model among already classified images [21].

Another paper [22] presents a method to improve content-based image retrieval with long-term relevance feedback. The method involves creating a sub-graph using fixed anchors and developing an efficient framework to predict ranking scores. User feedback is utilized to update the sub-graph and enhance retrieval performance.

In this paper, they utilize Manifold Ranking (MR) and Efficient Manifold Ranking (EMR) for CBIR. However, these methods are not suitable for large databases. Their proposed approach constructs a sub-graph based on fixed anchors, which simplifies the updating process. They also develop an efficient framework to predict ranking scores. To enhance the ranking system's performance in the long term, they incorporate long-term user relevance feedback by extending the queries and updating the weights in the graph [22].

*B. Short-term Learning*

The method proposed in [23], called "weight-learner," is an approach for content-based image retrieval that aims to bridge the gap between visual features and user semantics. It focuses on learning the best weights for visual features to connect low-level features and high-level user semantics. The method uses an optimization problem to compute feature weights based on the user query. It learns the optimal weights for each visual descriptor without supervision. These weights are obtained by minimizing an objective function that considers the distances between query images and images in the database. User input is not needed for feature weights or query image scores. This method can be applied to any visual feature such as color, texture, or shape. The feature weights are learned automatically from the set of query images without requiring additional information from the user. Experimental results show the advantages of the weight-learner approach compared to other systems for single-query retrieval and multiple-query retrieval.

A study [24] introduces a hybrid CBIR system improving bone tumor radiograph retrieval. It integrates quantitative and semantic features with user feedback, significantly boosting performance compared to the baseline system, with an average precision of about 0.90 versus 0.20. The approach overcomes CBIR limitations by merging relevance feedback and predicting semantic features, targeting enhanced efficiency and potential for combining semantic CBIR with relevance feedback in interpreting radiological images.

The system combines three types of information: image quantity-based features, semantic features, and user feedback. It operates in offline processing, building an annotated radiograph database, and online operation, retrieving similar images based on combined features and user feedback [24].

In the papers [13,25], the authors investigate the use of Convolutional Neural Networks (CNNs) in CBIR systems to improve relevance feedback. The authors in [13] compare two CNN architectures for image representation and RF, demonstrating the effectiveness of CNNs in extracting features for image retrieval tasks. The study recommends utilizing CNNs as feature extractors in image retrieval tasks without fine-tuning, as they yield effective and generalizable results. The authors propose two CNN architectures for relevance feedback: one modifies the last layer, while the other adds an extra layer. The study emphasizes the significance of robust feature selection and representation in CBIR systems. The proposed strategies involve using CNNs as feature extractors and classifiers to refine and reformulate queries based on user feedback [13].

The mentioned paper derives both CNN architectures from the well-known AlexNet model (refer to Fig. 1). Here's a brief explanation of each architecture:

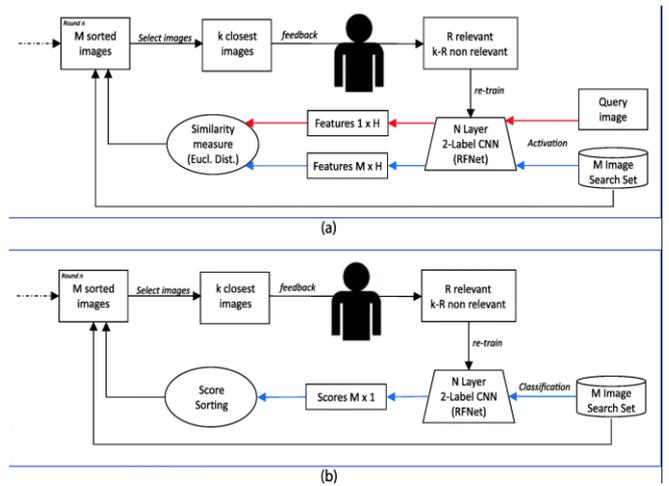

Fig. 1. (a) A visual retrieval system using a Convolutional Neural Network (CNN) for extracting features and (b) classifying images is depicted in the conceptual representation [13].

1. Architecture 1: This architecture maintains the original depth of the AlexNet model. Only the last layer of the network is modified to suit the relevance feedback (RF) task. The modified layer is fine-tuned for distinguishing relevant and non-relevant images based on user feedback. User-labeled images are utilized as the training set for retraining this layer.

2. Architecture 2: This architecture involves adding an extra layer to the original AlexNet model. Similar to Architecture 1, the last layer of the network is modified to handle the RF task. The modified layer is fine-tuned using user-labeled images as the training set.

Both architectures aim to extract features or classify images based on their relevance to enhance the image retrieval process. The paper discusses various strategies to improve the query refinement based on user feedback and achieve more precise retrieval outcomes [13].

In paper [26], a method for retrieving synthetic aperture radar (SAR) images using content, with initial retrieval and refined results with relevance feedback, was suggested. It introduces a new similarity measure called region-based fuzzy matching (RFM) and uses multiple relevance feedback (MRF) to enhance initial retrieval. MRF involves separate application of various feedback approaches, which are then combined to improve initial retrieval. The goal is to improve accuracy by including user feedback or specific algorithms to select relevant and irrelevant samples within the retrieved imagery.

In MRF, relevance feedback is applied iteratively using active learning (AL) algorithms to automatically select SAR images. AL algorithms iteratively choose SAR images and their binary labels (relevant and irrelevant) are used to retrain the binary support vector machine (SVM) classifier, which then re-ranks SAR images within the dataset [26].

MRF aims to intelligently refine initial retrieval results by using multiple relevance feedback approaches to iteratively enhance the ranking of SAR image patches. This iterative procedure continues until certain convergence conditions are met, aiming for greater precision in retrieval results [26].

The paper [27] suggests a method for content-based image retrieval using Particle Swarm Optimization (PSO) in relevance feedback-based systems. PSO is an algorithm inspired by the social behavior of bird flocking or fish schooling. It is used to find the optimal feature weights, allowing for more sophisticated, non-linear relationships between the features. User feedback on retrieved images guides the PSO optimization process. This method focuses on optimizing feature weights for color, shape, and texture to improve image retrieval accuracy. It uses user feedback to refine the retrieval process and reduce computational complexity.

To bridge the semantic gap and improve object detection, we have explored a short-term learning (STL) framework based on relevance feedback for content-based image retrieval [11]. It starts by finding images based on basic features, and then users identify some of these images as being related to their original query. Our method is inspired by the "Near Strangers or Distant Relatives" model, which uses the idea of family relationships. In this context, images in the same category are seen as relatives, while those in different categories are considered strangers. Using these user-labeled images, the method aims to gradually improve the relevance of the retrieved images.

Following the first round of finding and labeling images, the method then uses the labeled images as new query images to find new relevant images. This process selects the most similar images to the query using a similarity measure, as the result of the iteration. With each iteration, the goal is to make the retrieved images more closely match the relevant ones and move further away from the irrelevant ones [11]. You can see this process in Fig. 2.

By incorporating relevance feedback and the STL framework inspired by the near strangers or distant relatives model, the proposed method effectively addresses the semantic gap and enhances object detection capabilities in CBIR systems [11].

This research [28] proposes a method for improving the retrieval of medical images by bridging the gap between low-level image features and high-level semantic concepts. The method enhances the recall and precision of the retrieval process by using a voting technique to select the most effective similarity coefficient based on retrieved images.

First, the RGB color space of the medical images is converted to the HSV color space, which is considered more suitable for extracting color moment features. Then, eighteen features are extracted from the images using color moment and GLCM (grey-level co-occurrence matrix) texture extraction functions. These features capture the color, texture, and shape characteristics of the images. In the next step, eight common similarity coefficients are employed to calculate similarity scores between the query image and the database images. These coefficients measure the similarity based on different criteria. The relevance feedback step involves using the top images retrieved from a single query for each class as voting populations to select the most effective similarity coefficient. This selected coefficient is then used for the final search process. By leveraging the voting from the top retrieved images, the method enhances the effectiveness of medical image retrieval [28].

The research paper [29] proposes a method to improve content-based medical image retrieval (CBMIR) systems. The method introduces two expansion methods as relevance feedback to bridge the semantic gap. The first method calculates the mean values of features from top-ranked relevant images and formulates a new expanded query image based on those mean values. This expanded query image is then used for the final search process. The second method applies a feature selection algorithm to select important features from the top-ranked images. The dimensions of the query image and the entire image dataset are then reduced based on these selected features. A new expanded query image is formulated using these important features, and it is used for the final search process. Experiments conducted using two medical image datasets have shown that these methods significantly improve retrieval effectiveness compared to using only the original features.

Research [30] proposes the bag-of-words association mapping method for semantic retrieval of remote sensing images. It has four steps:

1. Extract features from smaller blocks of the image using techniques like Local Binary Pattern (LBP) and color moment analysis.

2. Create a visual feature word bag by combining feature vectors from multiple training set blocks. Each word is linked to its semantic set.

3. Mine association rules between visual feature words and semantics using an improved FP-Growth (frequent pattern tree) algorithm (an efficient and popular algorithm used for mining frequent itemsets and generating association rules in a dataset). The rules show connections between specific features and the semantics they represent.

4. Use a feedback mechanism to improve retrieval accuracy. Users can correct incorrect semantics by labeling images or indicating incorrect semantics. The system analyzes feedback and adjusts association rules to enhance future retrievals.

As a result, the bag-of-words association mapping method establishes a direct connection between visual features and semantics, enabling accurate retrieval of relevant remote-sensing images [30].

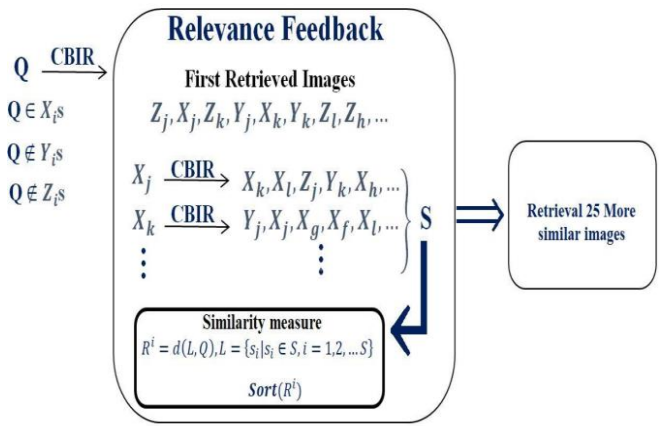

Fig. 2. Fig. 1. The proposed short-term learning method based on relevance feedback for CBIR in [11].

The paper [31] discusses the idea of "explainability" in CBIR systems and suggests using Similarity Based Saliency Maps (SBSM) to show which parts of an image contribute to the system's decisions. CBIR systems retrieve images using feature representations and similarity distances. However, these representations lack semantic interpretability, which makes it hard for users to understand how the system works.

To create SBSMs, a binary mask blocks out specific areas of a retrieved image. The importance of the blocked region is then estimated based on its impact on the similarity metric used in CBIR. By sliding the binary mask over the image, the method reveals the significance of different blocked areas. The SBSM method generates heatmaps that highlight the most relevant areas of an image (where "hotter" regions indicate higher relevance to the query) in CBIR [31].

This method incorporates relevance feedback, where users mark images as "relevant" or "not relevant" during the CBIR process. The feedback is collected through the Interactive Query Refinement (IQR) feature of the Social Media Query Toolkit (SMQTK). The combination of feedback and SBSMs helps users make informed decisions and gain insights into how the CBIR system operates. The correlation between user ratings (Likert scores) and retrieval accuracy demonstrates the usefulness of relevance feedback in improving search efficiency [31].

Another paper [32] presents a new way to retrieve patterned fabric images. It utilizes the non-subsampled contourlet transform (NSCT) feature descriptor along with the relevance feedback technique. This approach combines color information with the NSCT feature descriptor, enabling better retrieval results.

Retrieving fabric images is difficult due to their diverse and complex appearance. The proposed approach incorporates color information in the HSV color space, along with the NSCT feature descriptor. The relevance feedback (RF) technique improves image retrieval based on user feedback by considering geometric similarity to update parameters and weights associated with relevant and non-relevant feature vectors [32].

The RF technique employs geometric similarity to update parameters and weights using relevant and non-relevant feature points. It utilizes information from non-relevant images to enhance the similarity measure. The technique adjusts the weights assigned to relevant examples, giving more importance to relevant points that are far from non-relevant points while reducing the weight for relevant points that are close to non-relevant points. The optimization problem aims to minimize the overall distance between the query vector and relevant vectors. The proposed algorithm iteratively updates the weights and parameters of the similarity metric [32].

In this section, we looked into some new papers on RF in CBIR. Nowadays, like other literature on computer vision and object detection, there is a growing focus on deep learning, particularly CNNs for FR.

In conclusion, the quality and quantity of user feedback also impact the effectiveness of the RF process. Limited or inaccurate feedback can hinder the system's ability to learn and adapt, resulting in suboptimal results. Encouraging users to provide comprehensive and reliable feedback is crucial for successful RF implementation.

Another challenge lies in selecting the relevant features to incorporate into the adaptation process. Determining which visual characteristics carry more relevance and appropriately combining different feature spaces is complex. Feature selection methods and adaptive weighting schemes are employed to address this challenge and improve the overall accuracy of the RF approach.

Furthermore, the iterative nature of RF methods may impose additional computational costs and delay the search process. Conducting multiple feedback iterations to refine query results can be time-consuming, which may not align with user expectations for real-time responses. Balancing the need for accuracy with efficient computation is a significant challenge in RF implementation.

Short-term relevance feedback algorithms are indeed more commonly used in CBIR. These algorithms focus on immediate adjustments to the retrieval process based on user feedback, allowing users to refine search results for the current session quickly and iteratively.

Short-term relevance feedback is especially well-suited for real-time retrieval scenarios where users require quick and relevant results without relying on extensive feedback history.

Long-term relevance feedback algorithms can provide sustained improvements over time, but they generally require more extensive data collection and analysis, making them less practical for regular users in real-time retrieval scenarios.

III. CONCLUSIONS AND FUTURE RESEARCH

In this survey paper, we provided an overview of Content-Based Image Retrieval (CBIR) systems and Relevance Feedback (RF) in the field of computer vision. It highlights the importance of CBIR in retrieving images based on visual content, rather than relying solely on metadata. The paper addresses the challenges faced by CBIR systems, including the semantic gap and scalability issues.

To bridge the semantic gap, we emphasize the incorporation of RF into CBIR systems. RF allows users to provide feedback on retrieved images, refining the search results and improving relevance. The survey presents different approaches in RF, categorized into Short-Term Learning (STL) and Long-Term Learning (LTL), which enhance the learning capability and accuracy of CBIR over time.

Furthermore, we explore various techniques and algorithms employed in CBIR systems. These include feature extraction methods such as color, texture, and shape, as well as the utilization of machine learning algorithms and deep Convolutional Neural Networks (CNNs) for enhancing retrieval accuracy. The importance of user-centric approaches, scalability, and the evaluation of CBIR systems was also discussed.

Based on our research, future directions for CBIR and RF include:

1. Deep Learning and CNNs: Use deep learning, specifically CNNs, to improve CBIR systems.

2. Scalability and Efficiency: Develop efficient algorithms and indexing methods for large image databases.

3. Bridging the Semantic Gap: Find ways to connect low-level visual features and high-level concepts for better retrieval.

4. Human-Computer Interaction: Improve user interaction through intuitive interfaces and feedback mechanisms.

In summary, this survey paper provided a comprehensive overview of CBIR and RF, highlighting their significance in image retrieval based on visual content. It identified challenges faced by CBIR systems and proposed potential solutions. The inclusion of RF allows for user feedback, improving search results and reducing the semantic gap. The paper also suggested future research directions, such as exploring deep learning techniques, user-centric approaches, scalability, and bridging the semantic gap. Continued advancements in CBIR and RF will contribute to improved retrieval accuracy, usability, and effectiveness in various application areas.